\pdfoutput=1

\documentclass[11pt]{article}

\usepackage[final]{nlp4MusA}

\usepackage{times}
\usepackage{latexsym}
\usepackage{booktabs}
\usepackage{tabularx}
\usepackage{enumitem}

\usepackage[T1]{fontenc}

\usepackage[utf8]{inputenc}

\usepackage{microtype}

\usepackage{inconsolata}

\usepackage{graphicx}

\title{HumMusQA: A Human-written Music Understanding QA Benchmark Dataset}

\author{Benno Weck \\
  Universitat Pompeu Fabra \\
  \texttt{benno.weck01@estudiant.upf.edu} \\
  \And
  Pablo Puentes \\
  Universitat Autònoma de Barcelona \\
  \AND
  Andrea Poltronieri \\
  Universitat Pompeu Fabra \\
  \And
  Satyajeet Prabhu\\
  Universitat Pompeu Fabra \
  \And
  Dmitry Bogdanov\\
  Universitat Pompeu Fabra \\
  \texttt{dmitry.bogdanov@upf.edu} \\}

\def\numquestions{320}
\def\numtracks{108}

\begin{document}
\maketitle
\begin{abstract}
The evaluation of music understanding in Large Audio-Language Models (LALMs) requires a rigorously defined benchmark that truly tests whether models can perceive and interpret music, a standard that current data methodologies frequently fail to meet.
This paper introduces a meticulously structured approach to music evaluation, proposing a new dataset of \numquestions{} hand-written questions curated and validated by experts with musical training, arguing that such focused, manual curation is superior for probing complex audio comprehension.
To demonstrate the use of the dataset, we benchmark six state-of-the-art LALMs and additionally test their robustness to uni-modal shortcuts.

\end{abstract}

\section{Introduction}
The rapid progress of Large Language Models (LLMs) has catalysed the development of Large Audio-Language Models (LALMs), such as Audio Flamingo~\cite{ghosh_audio_2025, goel_audio_2025} and Qwen-Audio~\cite{chu_qwen-audio_2023}.
These multi-modal systems integrate an audio encoder with a large language model, allowing them to process audio input and generate textual responses conditioned on what they hear. 
This sets them apart from earlier self-supervised audio representation models \cite{schneider2019wav2vec, alonso2025omar}, which learn acoustic features without language generation, and from uni-modal text-only approaches.
To achieve comprehensive audio understanding, LALMs must go beyond speech recognition and encompass all audio domains \cite{iyer_analyzing_2025}, with music being one of the most challenging -- requiring a model to listen to an audio clip, process a text-based question, and produce an answer grounded in auditory perception.

Music understanding presents persistent challenges for LALMs due to music's dynamic, layered, and information-dense nature. 
This includes both perceptual and analytical capabilities, recognizing musical features like instrumentation, key, and structure, as well as cultural and contextual knowledge about genre and mood.
Evaluating music understanding in LALMs is particularly difficult because musical concepts are often complex and open-ended, making conventional lexical metrics like BLEU~\cite{papineni_bleu_2002} inadequate for assessing the diverse language responses.

To establish a comprehensive and objective measure of auditory intelligence, the field has coalesced around Question Answering (QA) frameworks~\cite[e.g.,][]{weck_muchomusic_2024,sakshi_mmau_2025,wang_audiobench_2025, yang_air-bench_2024}, 
which structure evaluation through multiple-choice classification, constrained reasoning, or open-ended questions that are better suited to assessing complex music capabilities.
Despite the growth of Music-QA datasets, the field has historically prioritized scale over quality.
Early benchmarks like MusicQA~\cite{liu_music_2024} and MusicInstruct~\cite{deng_musilingo_2024} were constructed by using LLMs to automatically augment existing captions from datasets like MusicCaps~\cite{agostinelli_musiclm_2023} or tags from MagnaTagATune~\cite{law_evaluation_2009}.
This reliance on automated sourcing often compromises evaluation integrity: text-only LLMs lacking audio perception can achieve high accuracy by exploiting language priors and ``world knowledge'' embedded in the question text alone.
This ``perception gap'' suggests that many current benchmarks primarily measure a model’s reasoning ability rather than genuine audio perception \cite{weck_muchomusic_2024,zang_are_2025}.

Automatically deriving questions from short, surface-level captions or tags inherently limits question depth and scope, 
preventing the formulation of challenging, multi-hop inquiries necessary for testing expert-level musical understanding.

Recent work has begun shifting toward expert-annotated benchmarks that demand more than surface-level recognition. 
A significant milestone in this direction is MMAU-Pro~\cite{kumar_mmau-pro_2025}, a comprehensive benchmark that utilizes expert-written and validated question-answer pairs to evaluate holistic auditory intelligence.
Notably, music forms a substantial portion of this dataset, with 1,618 questions dedicated to musical understanding.
While MMAU-Pro sets a high standard for expert curation, it highlights a remaining trade-off in benchmark construction regarding data provenance.
To avoid data leakage from existing training sets, MMAU-Pro sources its audio ``from the wild'' and through various online repositories.
This approach, while robust against leakage, often relies on disparate sources with potentially variable audio quality and metadata reliability.
Furthermore, other expert-curated efforts like MusicTheoryBench (MTB)~\cite{yuan-etal-2024-chatmusician} offer high expert-driven quality but remain limited to the symbolic domain (ABC notation), failing to test direct perceptual grounding in audio.

We argue that evaluating the full depth of music understanding requires a specialized, perceptually rigorous approach that combines expert curation with high-fidelity source material.
We introduce a novel evaluation dataset containing \numquestions{} hand-written questions, curated and validated by experts with advanced musical training.
Manual authorship enables broader topic coverage and more sophisticated multi-layered reasoning than automated generation can achieve. 
Crucially, our design minimizes language shortcuts: questions require genuine musical perception and analysis across structural, harmonic, perceptual, and cultural dimensions. 
All audio materials are sourced from Creative Commons-licensed recordings, ensuring the benchmark can be openly distributed.

\section{Methodology}

The goal of this study is to create a human-authored benchmark for evaluating large audio-language models on music understanding tasks.
The benchmark consists of \numquestions{} expert-written questions paired with freely licensed musical recordings, designed to assess model performance across diverse aspects of musical knowledge and reasoning.

All audio tracks were sourced from Jamendo\footnote{\url{https://www.jamendo.com/}}, a platform hosting Creative Commons-licensed music. 
We selected \numtracks{} tracks spanning multiple genres, instrumentation types, and production styles to ensure comprehensive coverage of musical characteristics. 
Each question refers to a specific excerpt from a track, ranging from $30$ to $90$ seconds in duration, with the exact time window determined by the question authors based on the musical content being assessed.
The use of openly licensed material ensures the benchmark can be freely distributed and reproduced without legal restrictions~\cite{bogdanov_mtg-jamendo_2019,manco_song_2023}, addressing a significant barrier to reproducibility in music AI research~\cite{batlle-roca_musgo_2025}.
The complete dataset, including questions, and metadata statistics is made publicly available under a Creative Commons licenses on Zenodo.\footnote{\url{https://doi.org/10.5281/zenodo.18462524}}

\subsection{Question Design}
\label{ssec:methodology/question-design}

Question design was informed by two established music education standards: the Associated Board of the Royal Schools of Music (ABRSM)\footnote{\url{https://www.abrsm.org/}} syllabi and the General Certificate of Secondary Education (GCSE) music curriculum\footnote{\url{https://www.gov.uk/education}}. 
We additionally informed our approach by existing music understanding benchmarks, such as MuChoMusic~\cite{weck_muchomusic_2024} and MMAU~\cite{sakshi_mmau_2025}.
Drawing on these sources, we designed questions spanning a broad spectrum of music understanding: from foundational perceptual tasks and world-knowledge aspects (e.g., cultural context, lyrical content) accessible to music beginners, to sophisticated analytical reasoning requiring music theory knowledge.

Three music theory experts (mean professional experience in music = 15 years; all holding advanced academic qualifications in music theory)  each authored approximately one-third of the questions. 
Authors were instructed to design questions that: 
i) reflect authentic educational objectives from ABRSM/GCSE curricula; 
ii) span diverse cognitive demands including perceptual identification, analytical reasoning, and interpretive assessment;
iii) require careful listening and musical knowledge to answer correctly; and 
iv) admit exactly one clear, unambiguous correct answer. 
All questions were designed in multiple-choice format with four options (one correct, three distractors) to facilitate automated evaluation.
Questions range from those accessible to casual listeners (e.g., \textit{``What emotion is mainly conveyed in this song?''} with options: joy, sadness, anger, disgust) to those requiring music theory knowledge (e.g., \textit{``What intervals create dissonance in the background guitar?''} with options: 4ths, fifths, octaves, unison).
More examples are provided in Appendix~\ref{sec:appendix}.

While LLMs could potentially generate questions of this kind, they lack the ability to ground questions in genuine audio perception. Expert authorship ensures that the proposed questions reflect authentic musical reasoning, requiring engagement with both the audio and the textual content. 

Experts played a dual role in this process: not only generating the questions, but also validating each other's work through iterative peer review. 
Each expert was asked to blindly answer questions authored by the others without prior knowledge of the intended correct answer, ensuring that questions could be consistently and unambiguously resolved. 
During this blind review, annotators flagged disagreements regarding the most likely answer option and provided written comments identifying potential issues. 
Common failure modes included questions deemed too subjective (e.g., relying on personal interpretation rather than objective musical features), distractors that were not equally plausible (e.g., one option being trivially eliminable), and incorrect or imprecise labeling of answer options. 
Authors then revised their questions based on this feedback, addressing flagged issues and clarifying ambiguities. This iterative cycle continued until no further comments or disagreements were raised, at which point the question was considered validated and included in the final benchmark. 

\subsection{Question Labelling}
\label{ssec:methodology/question-labeling}

To enable systematic analysis of model performance across different aspects of musical understanding, we classified questions according to two dimensions: \textit{musical category} and \textit{level of musical knowledge required}.

Each question was assigned one or more categories from an adapted version of the MuChoMusic~\cite{weck_muchomusic_2024} taxonomy, comprising 13 musical dimensions: Melody, Harmony, Metre and Rhythm, Instrumentation, Musical Texture, Sound Texture, Performance, Structure, Mood and Expression, Lyrics, Genre and Style, Historical and Cultural Context, and Functional Context. 
Each question received one primary category reflecting its main analytical focus and zero or more secondary categories if addressing multiple aspects.

We additionally classified questions according to the level of musical knowledge required to answer them correctly based on listening alone. 
Questions were assigned to one of three levels: Low (answerable by casual listeners with no formal training), Medium (requiring some musical training or active listening experience), or High (requiring formal music education or specialized knowledge).

Both classifications were performed using GPT-5 (\texttt{gpt-5-2025-08-07}) \cite{openai_gpt-5_2025} with structured prompts providing category definitions and detailed examples of each dimension. 
We used LLM-assisted annotation to ensure consistency across all 320 questions and 13 categories, reducing subjective interpretation of category boundaries. 
Two domain experts independently validated all automated assignments and disagreements were resolved through discussion. 

The final benchmark comprises questions distributed across all 13 musical categories, with the most frequent being Instrumentation (19.7\%), Harmony (11.3\%), and Melody (10.6\%), while Musical Texture, Structure, and Lyrics each represent less than 3.5\% of questions. 
Regarding difficulty, 44.4\% of questions were classified as \textit{low}, 38.4\% as \textit{medium}, and 17.2\% as \textit{high}.

\begin{table*}[th]
\centering
\small
\begin{tabular}{@{}l cccc cccc@{}}
\toprule
 & \multicolumn{4}{c}{Accuracy with SD (\%)} & \multicolumn{4}{c}{Consistency (\%)} \\
\cmidrule(lr){2-5} \cmidrule(lr){6-9}
Model & All & Low & Med. & High & All & Low & Med. & High \\
\midrule
qwen2.5-omni-7b \cite{xu_qwen25-omni_2025}     & \textbf{64.3} \textpm{} 2.9 & \textbf{73.9} \textpm{} 3.1 & \textbf{58.1} \textpm{} 2.5 & 53.2 \textpm{} 3.9 & 66.6 & 70.5 & 61.3 & 67.9 \\
music-flamingo \cite{ghosh_music_2025}     & 58.5 \textpm{} 1.6 & 60.9 \textpm{} 2.1 & \textbf{58.1} \textpm{} 1.8 & 53.2 \textpm{} 4.1 & 39.1 & 41.7 & 38.7 & 30.2 \\
audio-flamingo-3 \cite{goel_audio_2025}   & 58.1 \textpm{} 3.2 & 62.7 \textpm{} 3.3 & 54.7 \textpm{} 3.8 & \textbf{54.1} \textpm{} 1.5 & 55.0 & 55.4 & 56.5 & 50.9 \\
gpt-audio \cite{openai_gpt-audio_2025} & 57.7 \textpm{} 1.5 & 65.3 \textpm{} 3.9 & 55.1 \textpm{} 1.9 & 43.6 \textpm{} 1.8 & 45.6 & 45.3 & 43.5 & 47.2 \\
gemini-2.5-flash \cite{google_gemini_2025}        & 55.9 \textpm{} 1.2 & 64.4 \textpm{} 1.5 & 48.4 \textpm{} 3.6 & 50.5 \textpm{} 2.7 & 35.0 & 40.3 & 31.5 & 28.3 \\
audsemthinker \cite{wijngaard_audsemthinker_2025}       & 54.8 \textpm{} 2.2 & 61.3 \textpm{} 1.5 & 49.0 \textpm{} 3.7 & 51.4 \textpm{} 3.5 & 42.5 & 44.6 & 41.1 & 39.6 \\ \addlinespace
gemini-2.5-flash \textit{(text-only)} & 42.7 \textpm{} 0.9 & 44.4 \textpm{} 0.5 & 37.6 \textpm{} 1.8 & 50.0 \textpm{} 2.0 & 48.8 & 54.0 & 46.0 & 43.4 \\
\bottomrule
\end{tabular}
\caption{Accuracy \& consistency scores for systems across all benchmark questions, overall and by difficulty level. Accuracy is averaged over four runs with randomized answer orderings (standard deviation shown). Consistency measures the percentage of questions where the model produced identical answers across all four runs, indicating robustness to answer position bias.}
\label{tab:acc_cons_scores}
\end{table*}

\section{Experiments}
To demonstrate the utility of our benchmark, we test several state-of-the-art LALMs, selecting models that span different design paradigms: general-purpose multi-modal LLMs (gemini-2.5-flash, gpt-audio), audio-specialized LALMs (audio-flamingo-3, qwen2.5-omni-7b, audsemthinker), and one model explicitly designed for music understanding (music-flamingo).
Since all models have been designed and fine-tuned on question-answering tasks, the QA format should be familiar, enabling performance to serve as a direct measure of music understanding rather than task format comprehension.
Furthermore, following prior work~\cite{zang_are_2025}, we assess whether questions can be answered using text alone, without access to audio.

\subsection{Evaluation strategy}

Models are evaluated by prompting them with audio snippets and corresponding multiple-choice questions.
Previous studies have shown that both large audio-language models and text-only large language models are highly sensitive to the ordering of multiple-choice options, with answer position alone inducing substantial performance variance and unstable model rankings \citep{lin_hearing_2025,zheng_large_2024,pezeshkpour_large_2024}.
To address this issue, and following established practices in recent audio and music understanding benchmarks \citep{weck_muchomusic_2024,lin_hearing_2025}, we evaluate each model under multiple randomized answer orderings.
Specifically, for each question, we perform four independent evaluation runs, where the answer options are randomly shuffled in each run.
Final performance metrics are computed by averaging results across these runs.

The output provided by the model is automatically parsed by an LLM (\texttt{gemini-2.5-flash}) prompted to match the response with the given options.
This ensures consistent analysis of model outputs of different lengths, particularly when responses are long.
From this matching, we calculate a simple accuracy scores which are presented in Table \ref{tab:acc_cons_scores}.

From the results, we observe that Qwen2.5-Omni-7B \cite{xu_qwen25-omni_2025} attains the best performance overall, as well as within the \textit{low} and \textit{medium} difficulty categories. 
Notably, this model demonstrates remarkable consistency, producing the same answers across multiple runs despite variations in answer option ordering. 
In contrast, most models exhibit strong sensitivity to answer shuffling, with performance varying substantially across runs -- suggesting vulnerability to prompt formulation rather than robust understanding.

\begin{figure}[th]
    \centering
    \includegraphics[width=1.0\linewidth]{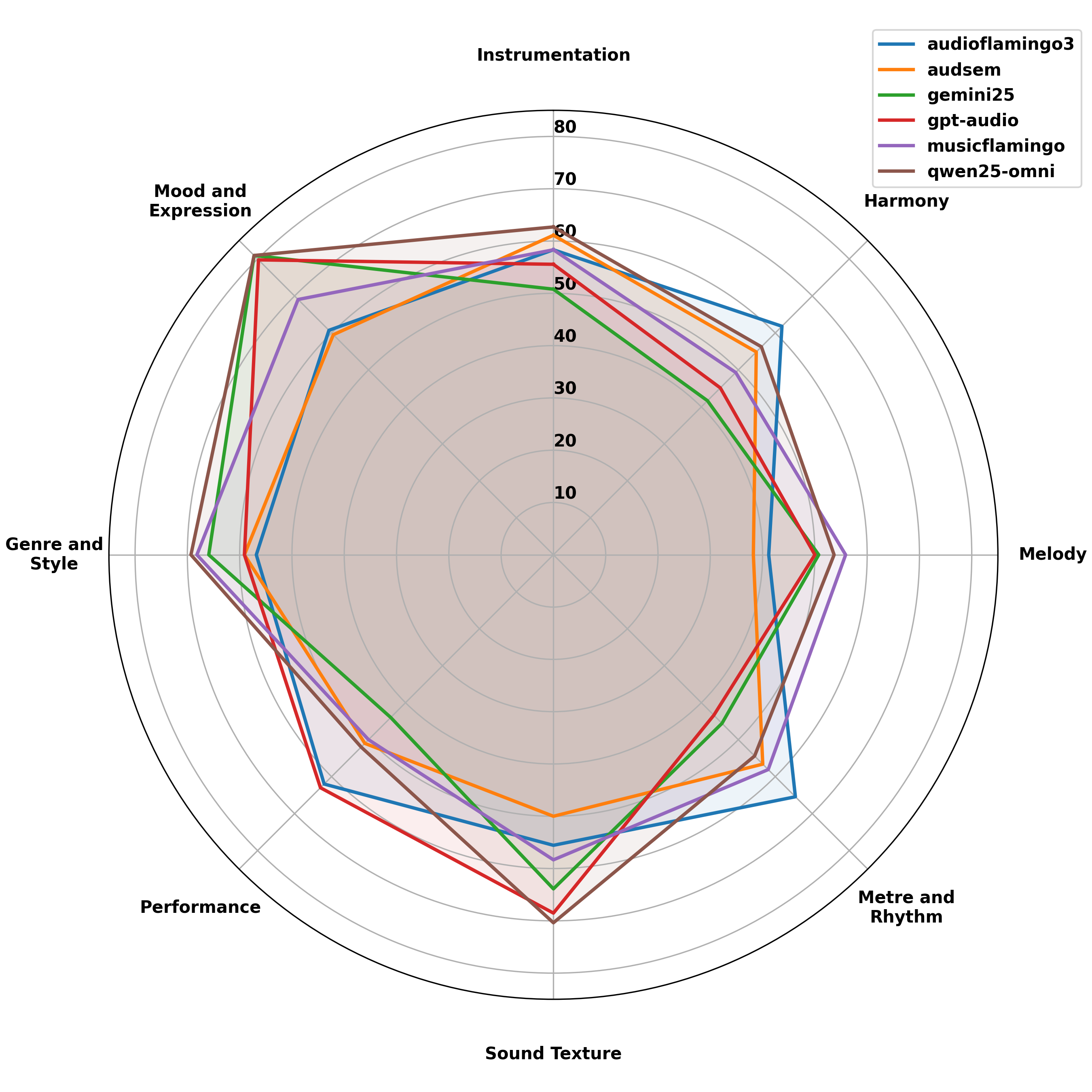}
    \caption{Accuracy across different categories in the benchmark. Categories accounting for $<=5\%$ of questions in the dataset are excluded from the chart. Specifically, these are \textit{Historical and Cultural Context}, \textit{Musical Texture}, \textit{Lyrics}, \textit{Structure}, and \textit{Functional Context}.}
    \label{fig:spider}
\end{figure}

Performance decreases consistently with increasing difficulty levels, validating our difficulty labeling scheme.
Figure~\ref{fig:spider} reveals distinct patterns across question categories: questions requiring general musical knowledge (e.g., genre and style, mood and expression, functional context) achieve higher scores, while music theory-grounded questions (e.g., harmony, melody, performance) yield substantially lower performance. 
This suggests that models are stronger at cultural and contextual reasoning than at formal analytical tasks requiring music theory expertise.

\subsection{Testing robustness to uni-modal shortcuts}

Additional experiments are conducted to test robustness by replacing real audio inputs with fake audio, following prior work \cite{zang_are_2025, weck_muchomusic_2024,kumar_mmau-pro_2025}.
The generally accepted hypothesis is that, without correct audio context, question-answering accuracy should not exceed random chance (25\%).
Therefore, we evaluate whether average model performance drops under these fake-audio conditions (see Table \ref{tab:fake_audio}).

\begin{table}[th!]
    \centering
    \small
    \begin{tabular}{lccc}
    \toprule
     & \multicolumn{3}{c}{Accuracy with SD (\%)} \\
    \cmidrule(lr){2-4}
    Model & Real & Noise & Silence \\
    \midrule
    qwen2.5-omni-7b & 64.3 \textpm{} 2.9 & 46.2 \textpm{} 1.3 & 47.0 \textpm{} 2.0 \\
    musicflamingo & 58.5 \textpm{} 1.6 & 42.9 \textpm{} 3.6 & 42.2 \textpm{} 3.0  \\
    gemini-2.5-flash  & 55.9 \textpm{} 1.2 & 37.8 \textpm{} 1.5 & 39.8 \textpm{} 1.6 \\
    \bottomrule
    \end{tabular}
    \caption{Accuracy scores with true audio compared to fake audio (Gaussian noise or silence).}
    \label{tab:fake_audio}
\end{table}

We also evaluate gemini-2.5-flash in a text-only setting, prompting it to respond using theoretical knowledge.
We further test using prompt variations and prompt optimisation strategies such as DSPy \cite{khattab_dspy_2024} without significant differences in results.
Results show that, while model performance is indeed degraded in comparison to the true audio setting, it exceeds pure random selection.
This is noteworthy since in our question writing we strive for equally likely options.

Analysis of responses reveals that the model exploits cues in the questions and employs its knowledge of common practices and methods like statistical likelihood of an option and elimination of outliers.
For example, in the question, ``The guitar is a typical accompaniment from a specific country. Which country is it?'', the model might deduce Brazil (Bossa Nova) by assuming ``typical'' means ``most globally recognized/distinctive,'' using theoretical and historical knowledge, and statistical prevalence of specific guitar styles.
Other options, Argentina, Venezuela and Cuba, while having ``typical'' guitar styles, are seen as less singularly iconic.
In another example, ``During the chorus, we can hear a very popular type of synthesizer sound. Can you guess its name?'', the phrase ``very popular type of synthesizer sound'' strongly cues ``supersaw'' over the other options (square, triangle, sine) because it's a named, popular sound, while the others are building blocks.
The options themselves create a categorical distinction.
These findings reveal that human-written questions are vulnerable to being answered due to factors such as weak question phrasing and distractors.
We leave a more detailed analysis of what makes human-authored questions solvable from text alone to future work.
In particular, follow-up studies could analyse why items labelled as high difficulty are still answered correctly by a text-only model in roughly 50\% of cases.
This could inform how annotation procedures might be refined to help annotators design genuinely high-difficulty questions (typically involving very specific conceptual content) such that all answer choices are roughly equally plausible based on the text alone.

\section{Conclusion}
We presented a music question-answering benchmark with \numquestions{} expert-authored questions and evaluated six state-of-the-art LALMs.
Results show that while models achieve moderate performance overall, they exhibit systematic weaknesses on music theory questions requiring analytical reasoning, with performance decreasing with difficulty.
The dataset is situated within the broader ecosystem of existing benchmark datasets and is intended to be used either in conjunction with or as a complement to them. Furthermore, it enables reproducible evaluation under clear licensing constraints, since both the audio materials and the expert-authored question text are provided under Creative Commons licenses.

\section{Acknowledgements}
We thank Armando Cedillo Mart\'{i}nez for the contributions and discussions.
This work is supported by “IA y Música: Cátedra en Inteligencia Artificial y Música” (TSI-100929-2023-1) funded by the Secretaría de Estado de Digitalización e Inteligencia Artificial and the European Union-Next Generation EU, under the program Cátedras ENIA.

\bibliography{references}

\twocolumn[
\clearpage
\section{Appendix}
\label{sec:appendix}
\appendix


\centering
\scriptsize
\begin{tabularx}{\textwidth}{p{1.65cm}lXX}
\toprule
Category & Difficulty & Question & Answer options (correct first) \\
\midrule

Functional Context & Low &
Where would you imagine this song being played? &
A: In a night club, B: At a children's party, C: During a romantic dinner, D: After exercising \\

 & Medium &
Which of these options is most related to this song? &
A: bedroom producer, B: racial protests, C: avant-garde music, D: futurism \\

Genre and Style & Low &
Which guitar comping style can we here in the tune? &
A: bossa nova, B: manouche guitar, C: flamenco guitar, D: hard bop \\

& Medium &
This song could have been written recently. However, its style sounds like it was written... &
A: Before bebop, B: After bebop, C: Before ragtime, D: After jazz fusion \\

 & High &
Right before the first verse, there are chords that are reminiscent of a particular jazz style. Do you know which one? &
A: Modal jazz, B: Be bop, C: Hard bop, D: Fusion jazz \\
\addlinespace
Harmony & Low &
This verse of the tune is written... &
A: In a major key, B: In a minor key, C: Neither of the two, D: It is not a tonal tune \\

 & Medium &
How would you define the harmonic rhythm of this tune? &
A: Steady, B: unstable, C: hemiolic, D: chaotic \\

 & High &
What surprises you about the harmony in this song? &
A: It has many second dominants, B: It uses chords with tensions, C: It uses polychords, D: Nothing is surprising \\
\addlinespace
Historical and Cultural Context & Low &
In what decade could this song be created? &
A: 2010, B: 1990, C: 1980, D: 1970 \\

 & Medium &
Where could the roots of this song be? &
A: Andalucia, B: Mexico, C: Cuba, D: Barcelona \\
\addlinespace
Instrumentation & Low &
What instrument plays the lead sound in the introduction? &
A: Synthesizer, B: Bass guitar, C: Piano, D: Ukulele \\

 & Medium &
The guitar is a typical accompaniment from a specific country. Which country is it? &
A: Brazil, B: Cuba, C: Argentina, D: Venezuela \\

 & High &
Which instrument plays the lead sound in the introduction? &
A: Muted trombone, B: Trumpet, C: Tuba, D: Saxophone \\
\addlinespace
Lyrics & Low &
What type of language is used in the lyrics of the song? &
A: Metaphoric and symbolic, B: Realistic, C: Narrative, D: Essay \\
\addlinespace
Melody & Low &
Which of the following best fits the guitar pattern heard at the beginning of this excerpt? &
A: riff, B: glissando, C: Alberti, D: walking bass \\

 & Medium &
During the introduction, the bass and the lead play their melodies in a... &
A: parallel motion, B: contrary motion, C: oblique, D: random \\

 & High &
In the verse there is a dissonance that doesn't belong the key and mode of the tune. Do you know which one is it? &
A: Ab, B: G, C: F, D: B \\
\addlinespace
Metre and Rhythm & Low &
What type of beat do you hear in the drums? &
A: rock, B: funk, C: soul, D: jazz \\

 & Medium &
During the introduction, the kick is played twice in each bar. On which beats? &
A: one and between the second and the third beat, B: two and four, C: one and three, D: two and three \\

 & High &
When the drums start playing, we can hear something known as... &
A: metric modulation, B: time signature change, C: swing, D: double-time \\
\addlinespace
Mood and Expression & Low &
What is the singer pursuing with her tone? &
A: Sensuality, B: Power, C: Loudness, D: Detachment \\

 & Medium &
What does not contribute to creating this mood? &
A: Tremolo, B: Reverb, C: The sound of the guitar, D: A clear and consonant harmony \\
\addlinespace
Musical Texture & Low &
What is the intention behind adding new voices in the last section? &
A: Increase energy, B: Decrease the energy, C: Maintain the same energy level, D: Introduce new energy \\

 & Medium &
What compositional technique do the piano and the drums use in the introduction? &
A: call and response (also known as question and answer), B: homophonically, C: There is only a piano in this excerpt., D: counterpoint \\
\addlinespace
Performance & Low &
What type of articulation best defines the melody of this song? &
A: legato, B: marcato, C: sforzando, D: staccato \\

 & Medium &
How does the drummer create anticipation for the final verse? &
A: Toms and snare fill, B: Cymbal fill, C: Hi Hat fill, D: Kick drum fill \\

 & High &
What extended technique is used by the flutist during the solo? &
A: Flutter tonguing, B: Shake, C: Spitting, D: Jet whistle \\
\addlinespace
Sound Texture & Low &
What effect is applied to the vocals in this track? &
A: pitch correction, B: Bitcrush, C: Ring modulation, D: gated reverb \\

 & Medium &
During the chorus, we can hear a very popular type of synthesizer sound. Can you guess its name? &
A: supersaw, B: square wave, C: triangle wave, D: sine wave \\

 & High &
What would you add to the drums to make them more present? &
A: A compressor, B: An equalizer, C: I would duplicate the track, D: I would play it in mono \\
\addlinespace
Structure & Medium &
What is the longest section of the tune? &
A: Solos, B: Verse, C: Chorus, D: Introduction \\

 & High &
Which section of the song follows this harmonic progression: vi, IV, I, V? &
A: chorus, B: verse, C: bridge, D: outro \\

\bottomrule
\end{tabularx}

\captionof{table}{Representative questions by category and difficulty}
\label{tab:examples}

]


\begin{table*}[ht]
\centering
\scriptsize
\setlength{\tabcolsep}{3pt}
\begin{tabularx}{\textwidth}{@{}
>{\raggedright\arraybackslash}p{1.6cm}  
>{\raggedright\arraybackslash}p{1.8cm}  
>{\raggedright\arraybackslash}p{1.8cm}  
>{\raggedright\arraybackslash}X         
>{\raggedright\arraybackslash}p{1.6cm}  
>{\raggedright\arraybackslash}p{1.2cm}  
>{\raggedright\arraybackslash}X         
>{\raggedright\arraybackslash}p{1.0cm}  
@{}}
\toprule
Model &
Backbone \& Architecture &
Training data &
Reasoning Mechanism \& Variants &
Context Length &
Max Audio &
Capabilities &
Openness \\
\midrule

audio-flamingo-3 &
Qwen-2.5-7B with AF-Whisper unified audio encoder &
AudioSkills-XL, LongAudio-XL, AF-Think, AF-Chat ($\sim$9.5M samples). &
\begin{minipage}[t]{\linewidth}\raggedright
Optional short chain-of-thought (on-demand, not default)
\begin{itemize}[nosep, leftmargin=*]
  \item \textbf{AF3 (Base)} - Single-turn Inference
  \item \textbf{AF3 + Think} - Inference with short CoT prefixes 
  \item \textbf{AF3-Chat} - multi-turn, multi-audio, voice-to-voice interaction
  \item \textbf{AF3 (Long-Audio reasoning)} - long-audio reasoning
\end{itemize}
\end{minipage} &
8{,}192 tokens &
Up to 10 minutes &
automatic speech recognition, audio question answering, audio reasoning, sound event recognition, music understanding, long-audio understanding, multi-turn audio dialogue, voice-to-voice interaction &
Fully open (research) \\
\midrule

audsemthinker &
Qwen2.5-Omni-7B fine-tuned &
AUDSEM dataset ($\sim$900k instances) from YouTube. &
\begin{minipage}[t]{\linewidth}\raggedright
Explicit structured semantic reasoning over sound
\begin{itemize}[nosep, leftmargin=*]
  \item \textbf{Think} - \texttt{<think> <answer>}
  \item \textbf{Think + Semantic} - \texttt{<think> <semantic\_elements> <answer>}
\end{itemize}
\end{minipage} &
32{,}768 tokens (inherited from Qwen2.5-Omni) &
Variable-length audio segments ($>$3 seconds, context-limited) &
semantic audio reasoning, sound event understanding, audio question answering, audio captioning, audio-based multiple-choice reasoning, creative audio-conditioned text generation &
Open \\
\midrule

gemini-2.5-flash &
Sparse Mixture-of-Experts (MoE) multimodal transformer &
Proprietary Data &
\begin{minipage}[t]{\linewidth}\raggedright
Dynamic internal reasoning with user-controllable thinking budget
\end{minipage} &
$\sim$1{,}000{,}000 tokens &
Up to $\sim$3 hours of audio/video &
automatic speech recognition, speech translation, audio question answering, audio summarization, audio-visual reasoning, long-context audio understanding, native audio dialogue, text-to-speech &
Closed / Proprietary \\
\midrule

gpt-audio-2025-08-28 &
Not available &
Not available &
Not available &
128{,}000 tokens &
Not available &
general audio model; accepts audio input and output; &
Closed \\

\midrule

music-flamingo-hf &
Enhanced AF3 backbone with time-aware embeddings (RoTE) &
MF-Skills (4M+) and MF-Think (300k) music datasets. &
\begin{minipage}[t]{\linewidth}\raggedright
Explicit music-theory chain-of-thought (optional, structured)
\end{minipage} &
$\sim$24{,}000 tokens &
Up to $\sim$20 minutes &
music captioning, music question answering, music theory reasoning, harmonic analysis, structural music analysis, lyric grounding, cultural music analysis, long-form song understanding &
Fully open (research) \\
\midrule

qwen2.5-omni-7b &
Qwen-2.5-7B Thinker-Talker end-to-end multimodal architecture &
Large-scale multimodal pretraining ($\sim$1.2T tokens) from open sources. &
\begin{minipage}[t]{\linewidth}\raggedright
Implicit internal reasoning
\end{minipage} &
32{,}768 tokens &
Long streaming audio (context-limited, no fixed minute cap) &
automatic speech recognition, speech-to-text translation, audio question answering, audio reasoning, sound classification, music classification, audio-visual reasoning, streaming speech interaction &
Open weights \\
\bottomrule
\end{tabularx}
\caption{Comparison of models used in the study}
\label{tab:audio_model_variants}
\end{table*}


\onecolumn
\section*{Summary of model-wise prompts and settings}

This section details the salient configuration parameters and representative prompt templates employed for each model evaluated in this study.

\bigskip

\begin{footnotesize}
\begin{enumerate}[nosep]
    \item \textbf{audio-flamingo-3}
\begin{verbatim}
model variant = 'Single-Turn Inference'
prompt = f"{question}\n{answer_str}"
\end{verbatim}

    \item \textbf{audsemthinker}
\begin{verbatim}
model variant = 'Think + Semantics' 
prompt = f"{question}\n{answer_str}"
\end{verbatim}

    \item \textbf{gemini-2.5-flash}
\begin{verbatim}
prompt = f"""**Task:** You are an expert musicologist with perfect pitch and extensive knowledge 
of music theory, instrumentation, and performance techniques. Your goal is to analyze the provided 
audio excerpt and answer the multiple-choice question with high precision. If audio is missing, 
use theoretical knowledge to deduce the answer.
{question}
{answer_str}
**Final Answer:** Return ONLY the single letter: A, B, C, or D"""
\end{verbatim}

    \item \textbf{gpt-audio}
\begin{verbatim}
prompt = f"{question}\n{answer_str}"
\end{verbatim}

    \item \textbf{music-flamingo}
\begin{verbatim}
prompt = f"{question}\n{answer_str}"
\end{verbatim}

    \item \textbf{qwen2.5-omni-7b}
\medskip   

Using a direct question-answer prompt did not yield optimal results. The model returns responses such as: 
\begin{verbatim}
"I'm not sure which direction the low - pass filter is shifting. It could be up or down, or even up - down. 
You might need to check the audio more closely or have some technical knowledge about filters to figure it 
out. Why are you interested in this low - pass filter?"
\end{verbatim}

We use a system prompt based on the audio understanding cookbook provided by the authors.\\
\url{https://github.com/QwenLM/Qwen2.5-Omni/blob/main/cookbooks/universal_audio_understanding.ipynb}\\
We further guide the model towards selecting one of the provided options using the following prompt template:

\begin{verbatim}

prompt = f"""You are a music audio understanding model.

Listen carefully to the provided audio clip. Answer the following multiple-choice 
question based on what you hear.

Question:
{question}

Options:
{answer_str}

Respond with ONLY the letter of the correct option (A, B, C, or D).
Do not include any explanation or additional text."""
\end{verbatim}
\end{enumerate}
\end{footnotesize}

\end{document}